# Design of a sensor integrated adaptive gripper

IA Sainul
Advanced Technology Development Centre, IIT Kharagpur,
Kharagpur -721302, India
sainul@iitkgp.ac.in

Sankha Deb
Mechanical Engineering Department, IIT Kharagpur,
Kharagpur -721302, India
sankha.deb@mech.iitkgp.ac.in

AK Deb
Electrical Engineering Department, IIT Kharagpur,
Kharagpur -721302, India
alokkanti@ee.iitkgp.ac.in

*Abstract*—In this paper, design and development of a sensor integrated adaptive gripper is presented. Adaptive grippers are useful for grasping objects of varied geometric shapes by wrapping fingers around the object. The finger closing sequence in adaptive grippers may lead to ejection of the object from the gripper due to any unbalanced grasping force and such grasp failure is common for lightweight objects. Designing of the proposed gripper is focused on ensuring stable grasp on a wide variety of objects, especially, for lightweight objects (e.g., empty plastic bottles). The actuation mechanism is based on movable pulleys and tendon wires which ensure that once a link stops moving, the other links continue to move and wrap around the object. Further, optimisation is used to improve the design of the adaptive gripper. Finally, validation is done by executing object grasping in a simulated environment PyBullet for simple pick and place tasks on common household objects.

*Keywords*—*Robot hands/grippers, Tendon-driven mechanisms Underactuation, Stable grasp, Impedance control*

## I. INTRODUCTION

Most of the industrial robots use simple two or three-finger grippers which are based on mechanisms such as linkages, gear trains, rack-pinions, etc. suitable for grasping objects of a particular geometric shape like cuboidal, cylindrical or spherical. On the other hand, service robots use dexterous multi-finger hands/grippers for performing more complex grasping and manipulation tasks. Underactuation mechanisms are used to reduce the complexity of such systems which need lesser number of actuators than the total degrees of freedom actually required. Such power transmission mechanisms must be adaptive (i.e., the gripper wraps/envelops its fingers around the irregularly shaped object) and the grasping force should be well distributed around the object for maintaining a stable grasp. Different variations of differential mechanism can be used to design such adaptive grippers [1], e.g., movable pulley [2], planetary or bevel gear differentials [3], seesaw mechanisms such as differential lever [4] and equalizing bar [5], etc. Other than the differential mechanisms, based on the requirements underactuated grippers can also be realised using linkages (high grasping force) or tendon wires (compact design, compliance). A number of robotic hands/grippers can be found in the literature [6]. All these robotic hands/grippers are either underactuated by means of tendon-pulley/linkage mechanism, or fully actuated by the same number of actuators as the number of DOF's. The Barrett hand is a commercially available three-finger underactuated gripper. The gripper uses a novel breakaway transmission mechanism [7] for realising the adaptive/ enveloping grasp. Masa et al. [8] developed an underactuated prosthetic hand based on the soft gripper mechanism [9] consists of pulleys and cable wires. A self-adaptive mechanism based on cable transmission and compression springs was designed in [10]. A highly underactuated anthropomorphic robotic hand using a differential mechanism based on tendon-pulley was presented in [11]. The Velo gripper [12] is a two-finger adaptive gripper very effective at performing parallel, enveloping and fingertip grasps. An underactuated three-finger gripper using only a single tendon and spiral springs for each finger was proposed in [13]. Although the underactuated mechanism has several advantages in terms of compact design, smaller in size, these grippers encounter problems when handling lightweight objects (e.g., empty plastic bottles) during adaptive/envelope/power grasp where the object slips away from the gripper [14]. Keeping the limitations in mind, it is, therefore, necessary to develop simple mechanisms putting more emphasis on grasp stability.

In this paper, a simple adaptive actuation mechanism based on movable pulley and tendons has been designed. A two finger gripper using the proposed mechanism has been optimised and developed which can perform adaptive/enveloping as well as fingertip grasps. Further, Magnetic position sensors have been incorporated at each joint of the fingers for measuring joint displacements. Finally, the experiment has been performed on common objects.

The rest of the paper is organized as follows. Section II describes the proposed adaptive gripper and its actuation mechanism. Section III presents design optimisation of the adaptive gripper. Section IV shows the results. Finally, Section V gives the conclusion.

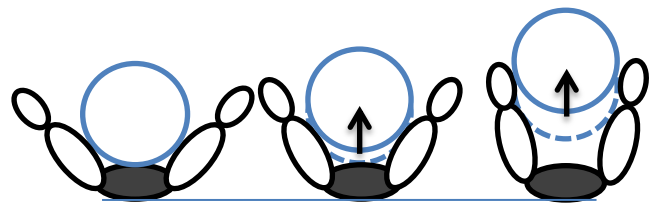

Fig. 1. (a) Fingers make contacts with the object (b) Object Splits out

## II. DESIGN OF THE GRIPPER

This work is mainly focused on designing a simple gripper mechanism, which can perform both enveloping/power and fingertips/precision types of grasps. Adaptive underactuated grippers adapt the shape of an object by wrapping the fingers around the object and usually have lesser number of actuators than the total number of degrees of freedom (dofs). The underactuated mechanism reduces the number of actuators which makes the design small, lightweight, and simple while keeping the gripper capability to adapt an object shape. A class of underactuated mechanisms [8, 9, 13, 14] are designed in such a way that the first link starts to move when actuation force applied and the subsequent links only move once the precedent link touches an object or maximum joint limit reached. Such mechanism encounters problems when handling lightweight objects (e.g., empty plastic bottles) during envelope/power grasp. Once the first links make contacts with the object, the resulting contact force push the object outward from the gripper and break the contacts. So, the first links move further to make contacts with the object at different location and the contact force further push the object outward. Meanwhile, the subsequent links do not move as the first links could not make stable contacts and the object slips away from the gripper. A solution to the object slipping away from the gripper can be simultaneously closing all the links and once a precedent link makes contacts with the object or maximum limit reached, the subsequent links continue to move.

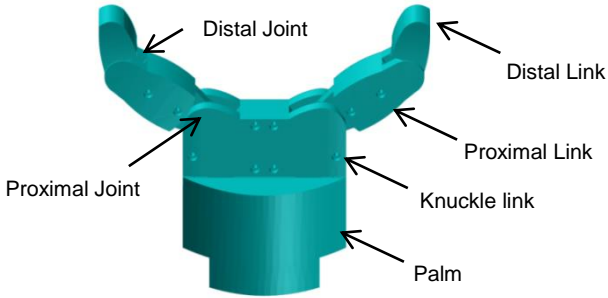

Fig. 2. Model of the robotic gripper

A model of the proposed robotic gripper is shown in figure 2. The gripper consists of two identical fingers each having three links namely knuckle, proximal, and distal phalanx and a palm. Knuckles are fixed on the palm and give support to an object during the enveloping/wrapping type of grasp. The knuckle-proximal and proximal-distal links are connected using two revolute joints to form an articulated finger. Four joints of the two fingers are actuated by a total of two DC motors. A tendon-spring system is used in each finger to achieve the flexion and extension motion of the middle and distal links of the finger.

*Underactuated finger mechanisms*

Keeping all the design requirements in mind, tendon driven mechanism is preferred over other mechanisms (e.g., linkages, gear train etc.) to achieve the desired finger motion. Tendon Driven Mechanism (TDM) offers advantages over other actuation mechanisms especially in terms of compact design and smaller in size of the gripper. In addition, the combination of tendons and springs exhibits inherent compliance which is desirable for a system interacting with the environment.

To achieve the desired finger motion, the actuation force is transferred from an actuator to the finger links through tendon wire. The tendon mechanism needs to distribute the actuation force to the links in such a fashion that the links start to move simultaneously and once a link touch an object or maximum limit reached the subsequent links continue to move. In order to make the mechanism work, a network of bifurcated tendons originated from the actuator with a combination of stretchable and non-stretchable tendon elements can be come up with certain advantages and disadvantages. A generalisation of the network with only non-stretchable tendons for n number of joints is shown in figure 3. The basic design of the network is that a single tendon originated from an actuator bifurcates into two branches at each joint of a finger. One branch is fixed with the immediate link of the joint and the other branch runs through the link, only to bifurcate at the next joint of the articulated finger.

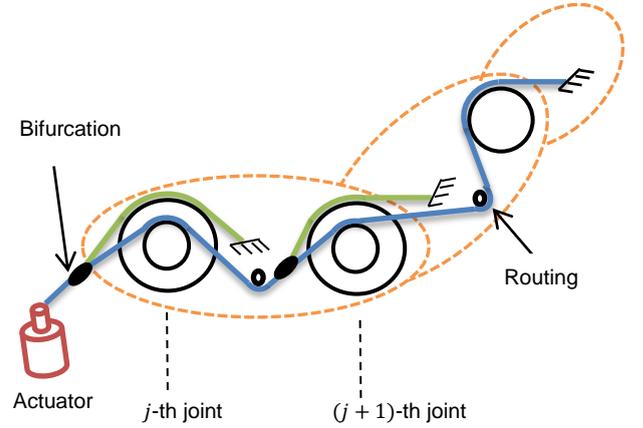

Fig. 3. Tendon network with non-stretchable tendons

Let n be the number of joints, $dl_j^1$ and $dl_j^2$ be the tendon displacements of the two branches at $j$-th joint, $dq_j$ be the joint velocity of $j$-th joint. For non-stretchable tendons, the displacements are equal for the tendon branches

$$dl_j^1 = dl_j^2 \tag{1}$$

For the $j$-th and $(j+1)$-th joints, the relation between the tendon displacement and the joint displacements is as follows.

$$dl_j^1 = r_j^1 dq_j \tag{2}$$

$$dl_j^2 = r_j^2 dq_j + r_{j+1}^1 dq_{j+1} \tag{3}$$

The equations (2) and (3) give the coupling relation between the joints as follows.

$$dq_{j+1} = \frac{(r_j^1 - r_j^2)}{r_{j+1}^1} dq_j \quad (4)$$

where, $r_j^1$ and $r_j^2$ are the pulley radii of the two tendon branches at $j$-th joint.

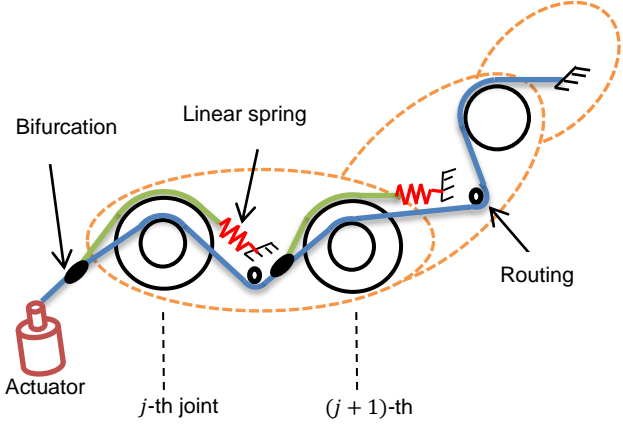

Fig. 4. Tendon network with spring loaded stretchable tendons

Now, when $j$-th link stops once it touches an object or maximum limit reached, i.e. $q_j = 0$, from equation (4) the $(j+1)$-th link also stops, which imply that the mechanism is underactuated but not adaptive. To make the mechanism adaptive, a stretchable tendon element is introduced in the first branch as shown figure 4. A linear spring at the end of the tendon can be added to make the tendon stretchable [15]. The modified relation (2) between the tendon displacement and the joint displacements for the stretchable spring-loaded tendon is as follows

$$dl_j^1 = \delta l_j^1 + r_j^1 dq_j \quad (5)$$

where, $\delta l_j^1$ is the elongation of the stretchable tendon. Then, equations (3) and (5) give.

$$dq_{j+1} = \frac{1}{r_{j+1}^1} \delta l_j^1 + \frac{(r_j^1 - r_j^2)}{r_{j+1}^1} dq_j \quad (6)$$

So, once a link stops, the subsequent likes continue to move and it depends on the elongation of the stretchable tendon.

Let $f_j^1$ and $f_j^2$ be the tendon forces of the two branches at the $j$-th joint, $\tau_j$ be the joint torque at the $j$-th joint. Now, the actuation force is equally divided at each bifurcation and the actuation torque generated by tendon branches depend on the moment arm of the associated joint pulley for each branch. Then, the tendon force at $j$-th bifurcation and the relation between the tendon force and joint torque is as follows.

$$f_j^1 = f_j^2 = f/2^j \quad (7)$$

$$\tau_j = r_j^1 f_j^1 + r_j^2 f_j^2 \quad (8)$$

where, f is the force generated by the actuator.

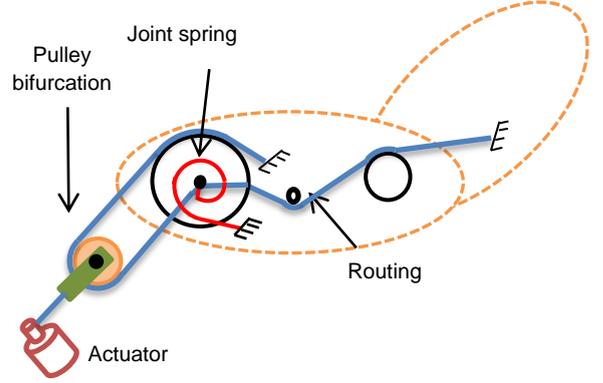

Fig. 5. Tendon network with pulley mechanism and non-stretchable tendons

Although the use of linear spring makes the design simple, the selection of the linear spring constant of the stretchable tendon is challenging. In this paper, an alternative mechanism is proposed to make the gripper adaptive and it works similar to the mechanism having stretchable tendons. The tendon network of the proposed mechanism consists of non-stretchable tendons and movable pulley mechanisms which are used for the tendon bifurcation as shown figure 5. The non-stretchable tendon originated from an actuator or the precedent bifurcation is fixed with the axle of the pulley and a non-stretchable tendon passes over the pulley whose one end is fixed on the immediate link and other end goes to the next pulley. Initially, the actuation force applied on the pulley produces an equal force on both sides of the tendon passing over the movable pulley. As a result, all the links move simultaneously until it touches an object or maximum limit reached. Once a link stops, the rotation of the movable-pulley facilities the subsequent links to continue its movement. The torque-force relation is decoupled by keeping the pulley $r_j^2 = 0$ as shown in figure 5. So, the relation between the tendon force and joint torque at $j$-th joint becomes.

$$\tau_j = r_j^1 f_j^1 \quad (9)$$

Initially, let the linear displacement of the bifurcation-pulley is $dl$ as shown figure 6, then the relation between the tendon displacements and the joint displacements are as follows.

$$dl_j^1 = dl = r_j^1 dq_j \quad (10)$$

$$dl_j^2 = dl = r_{j+1}^1 dq_{j+1} \quad (11)$$

Once a link touches an object or maximum limit reached, the relation between the tendon displacement and the joint displacements is as follows.

$$dl_j^1 = 0 \quad (12)$$

$$dl_j^2 = 2dl = r_{j+1}^1 dq_{j+1} \qquad (13)$$

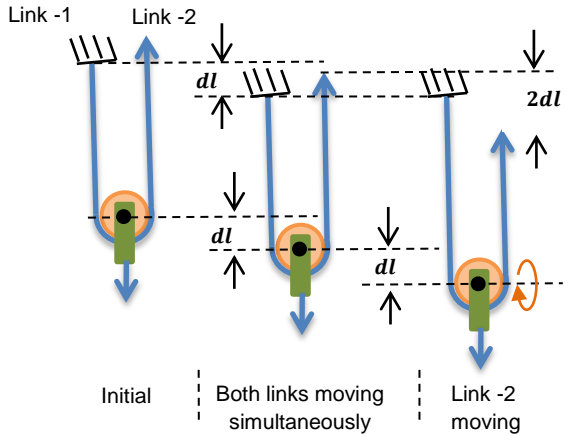

Fig. 6. Different stages of the movable pulley: initially the movable-pulley at equilibrium, both the links move simultaneously; once a link stops, the pulley starts to rotate which allows the other link to continue its movement.

The tendon-pulley system only produces flexion motion to the finger where extension motion is achieved using spiral springs at the joints. Each finger consists of three links namely knuckle, proximal and distal. The pulleys join them to form an articulated chain from knuckle to distal link. The choice of the pulley radius is important because it decides the essential torque that needs to be generated at the joint.

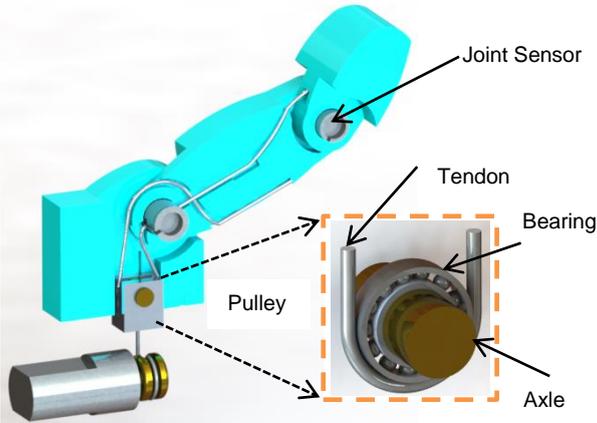

Fig. 7. Tendon network with spring loaded stretchable tendons

The bifurcation pulley is realized using small bearing supported by a small movable part. The movable part is connected to the actuator through a tendon wire and moves linearly along the slotted channel inside the knuckle and palm. Another tendon wire passes over the bearing and then the branches pass over the joint pulleys and run through the routing path inside the finger. The tendon path and actuation system are shown in figure 7.

### III. DESIGN OPTIMISATION

The main focus of the gripper design is to accomplish a stable grasp for a wide range of objects, especially for lightweight objects (e.g., empty plastic bottles) during envelope/power grasp. On the other hand, ensure that the gripper is not limited to only the enveloping type of grasp and it is equally effective for fingertip type of grasp. On this regards, the optimisation is used to improve the adaptive mechanism and dimension of the gripper.

#### A. Optimisation of link dimensions

It is desirable to choose the link dimension in such a way that the gripper able to wrap its fingers around the object during enveloping grasp, in other words, all the links make contacts with an object. Now it can be observed from figure 8, the choice of smaller links fails to fully wrap the object, while a collision between the fingers prevents the gripper making contact with the object due to choice of larger link dimensions.

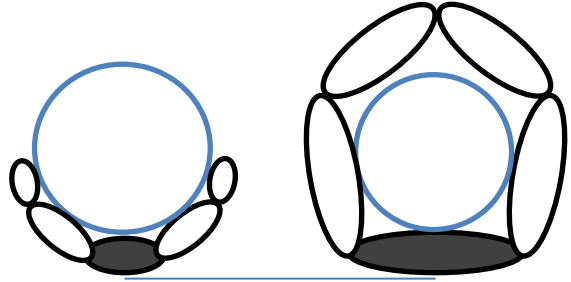

Fig. 8. (a) Fingers with smaller links fail to fully wrap the object, (b) fingers with larger links collide each other.

A brief formulation is given to find the grasp quality where more details can be best found in Miller and Allen [16]. The contact friction is modelled by approximating the friction cones with eight-sided pyramids having a unit length and a half angle of $tan^{-1}\mu_s$, where $\mu_s$ is the static friction coefficient as shown in figure 9. Then the contact force $\boldsymbol{f}$ transfer to the object through contact is the linear combination of the vectors used to approximate the eight sides of the pyramid.

$$\boldsymbol{f} = \sum_j^m \alpha_j \boldsymbol{f}_j \qquad (14)$$

where, $\alpha_j > 0$, $\sum_j^m \alpha_j = 1$ and number of sides of the pyramid $m = 8$

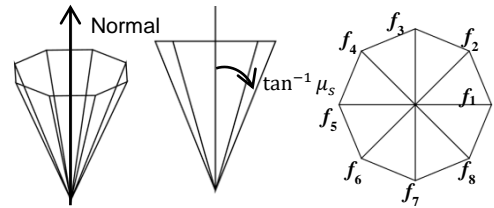

Fig. 9. Approximation of the friction cone with an eight-sided pyramid

The contact wrench can be defined as the six dimensional vectors formed by contact forces and torques at the contact point and is given as follows.

$$w_{i,j} = \begin{pmatrix} f_{i,j} \\ \lambda(d_i \times f_{i,j}) \end{pmatrix} \quad (15)$$

where, $f_{i,j}$ is the $j$-th component of the friction cone at the $i$-th point of contact. $d_i$ is the distance vector from torque origin to the $i$-th point of contact. The scalar $\lambda$ enforces the constraint $\|\tau\| \leq \|f\|$.

Now, assembling all the contact wrenches gives the convex hull or the polygon as follows.

$$W = ConvexHull\left(\cup_i^n \{w_{i,1}, w_{i,2}, \ldots, w_{i,m}\}\right) \quad (16)$$

The volume $v$ of the hull gives an invariant measure of grasp quality subjected to the origin of the wrench space lies within the convex hull. This quality measure is used as a basis for the optimisation of the link dimensions. The optimisation problem is formulated for the knuckle, the middle and the distal links of the proposed gripper. The cumulative grasp quality ($Q$) over a pool of valid grasps is maximised to find the best link dimensions. Where a pool of grasp is created from an object database consists of household objects of varied shapes and sizes.

$$Q = \sum_j^N v_j \quad (17)$$

where, $N$ is the size of the grasp pool.

*B. Optimisation of the actuation mechanism*

The main objective of gripper design is to provide a stable grasp by minimising unbalance grasping force on the object, which often leads to object ejection from the gripper. The link dimension optimisation only ensures that the gripper will able to fully wrap a wide range of objects, whereas the grasping force on the object can be regularized by optimising the actuation mechanism. The pulley radii and joint spring stiffness are the main design parameters which decide the behaviour of the mechanism and are considered for optimization. The best way to prevent object ejection from the gripper is to close the fingers in such a way that all the links make contact simultaneously as discussed earlier. Now, the pulley ratio decides the joint displacement of the distal to the proximal joint, in other words, how quickly the distal link closes relative to the proximal link. The choice of small pulley ratio results in an imbalance contact force as only proximal links can make contact with the object which leads to grasping failure, while large value makes the distal links touch the object without proximal link contacts which results in a weak grasp as shown in figure 10. So, it is important to choose the pulley ratio in such a way that it fits a wide range of objects. The idea is to use joint displacement data to estimate a model for the pulley radii ratio to avoid the two extreme cases and try to find configuration for a wide range of objects. Here, simple regression is used to estimate the pulley ratio over a pool of grasps consisting of a wide range of objects.

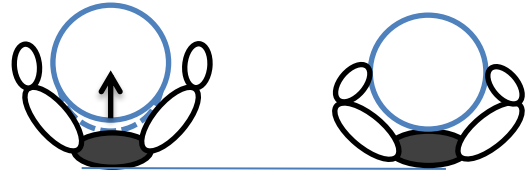

Fig. 10. Fingers make contacts with the object (left), quick closure of the distal link leaves no time for proximal links to make contacts (right)

IV. RESULTS AND DISCUSSIONS

The optimisation technique has been used to decide the design parameters of the adaptive gripper. The optimisation of the design parameters has been done over a dataset of total eighteen common household objects. Twelve objects are taken from the KIT database [17] and six objects are designed. The Simulated Annealing (SA) technique has been applied to optimise the link dimension of the fingers. The limits of the link length as given in table 1 are chosen based on the physical limitations and several trials of the optimization. The mean values of the link limits set the seed values of the optimisation variables and other parameters of the SA. In each iteration, the new solutions of the optimisation variables are generated using random Gaussian distribution. The pool of grasps on the object dataset is generated using the object slicing based grasp planner [18]. Then, the best link lengths are found by the maximising the cumulative grasp quality over the grasp pool. The optimisation converges in 200 iterations and the best lengths of the knuckle, proximal and distal links are 34.2mm, 58.4mm and 44.4mm, respectively.

The pulley ratio is estimated from the joint displacement data of the found grasp over the object dataset. The simple regression technique estimates the pulley ratio of 1.35 over the pool of grasps consisting of 760 grasps, where a subset of 6 grasps is shown in figure 11.

**Table 1** Lower and upper limits of the link lengths

|  | **Knuckle** | **Proximal** | **Distal** |
|---|---|---|---|
| Minimum (mm) | 25 | 45 | 25 |
| Maximum (mm) | 35 | 75 | 45 |

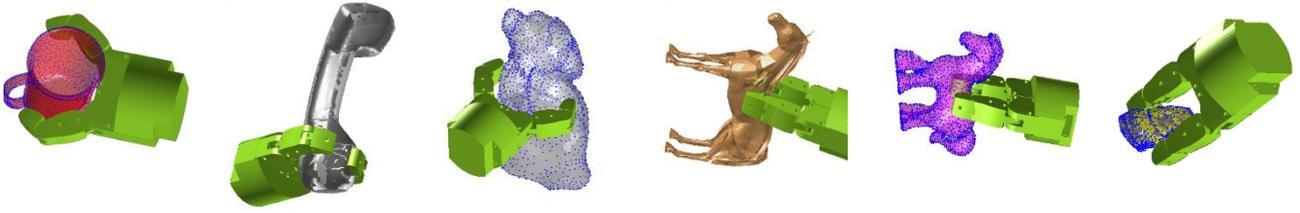

Fig. 11. A subset of the grasp pool found using the object slicing based grasp planner.

The physical prototype of the optimised gripper is manufactured using 3D printing technology as shown in figure 12. PLA material is used to print the gripper parts. The major challenges of choosing an appropriate motor for the finger actuation are the limited physical space available within the finger and the high required torque. Small dc motors are chosen, which can generate maximum rated output torque of 0.00569 Nm, maximum rated stall torque of 0.0125 Nm and can achieve a rotational speed of 1200rpm. The motors have a built-in gearbox with a gear ratio of 100:1, and a pulley is fixed with the output shaft, which drives the tendon. The total weight of the motor and gearbox is 50g. Two Small dc motors are placed inside the palm for pulling the tendons, one motor for each finger. A single motor controls two joints of a finger. Hall-effect based absolute position sensors are used at the joints, which measure the joint displacements of the fingers. The sensor has a small permanent magnet and a hall-effect ic. The ic is mounted on a small printed circuit board (PCB) and placed at each joint. The magnet is mounted on the shaft of the joint pulley.

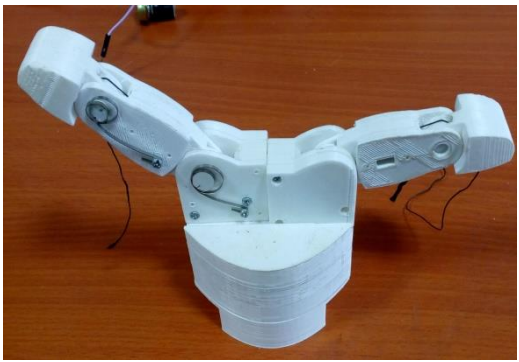

Fig. 12. A 3D printed prototype of the two fingers adaptive gripper

*Object grasping test*

The proposed mechanism is validated by executing object grasping in a simulated environment PyBullet [19]. The PyBullet is the python interface of the open physics engine Bullet. The gripper is mounted on an industrial Motoman manipulator for performing simple pick and place operation on the planned grasps. The examples of successful grasp by the proposed simultaneous wrapping of the finger links around the object and the mechanism, where the links are closing one-by-one, resulted in an unsuccessful grasp are shown in figure 13 (a) and (b), respectively.

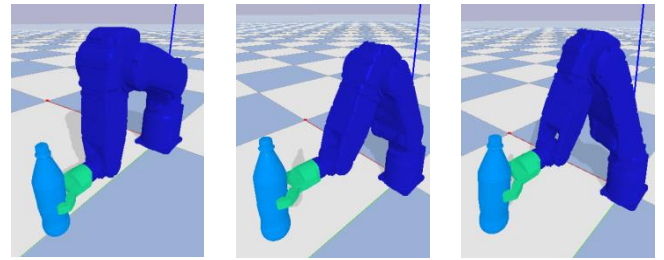

(a)

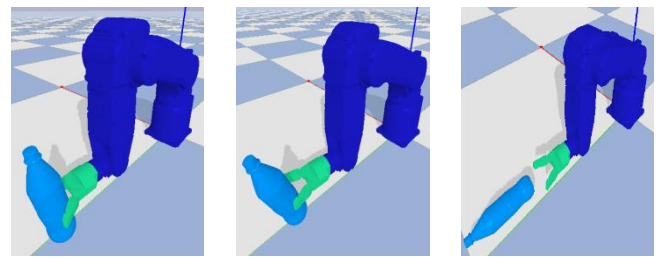

(b)

Fig. 13. Simple pick and place operation, (a) successful (b) unsuccessful.

## V. CONCLUSION

In this work, a simple adaptive actuation mechanism based on movable pulley and tendons has been designed. The optimisation of the design parameters has been done over a dataset of total eighteen common household objects. A two-finger gripper based on the optimisation results has been developed using 3D printing technology, which can perform adaptive/enveloping as well as fingertip grasps. Further, Magnetic position sensors are incorporated at each joint of the fingers for measuring joint displacements. Finally, the proposed mechanism is validated in a simulated environment PyBullet.